%% file: frstarcnn.tex
\documentclass[10pt,twocolumn,letterpaper]{article}

\usepackage{iccv}
\usepackage{times}
\usepackage{epsfig}
\usepackage{graphicx}
\usepackage{amsmath}
\usepackage{amssymb}
\usepackage[usenames,dvipsnames]{xcolor}
\usepackage{lib}

\usepackage[pagebackref=true,breaklinks=true,letterpaper=true,colorlinks,bookmarks=false]{hyperref}

\iccvfinalcopy 

\def\iccvPaperID{0750} 

\newcommand{\vecwp}{\bfw_\textrm{p}^{\alpha}}
\newcommand{\vecws}{\bfw_\textrm{s}^{\alpha}}

\ificcvfinal\fi
\begin{document}

\title{Contextual Action Recognition with R*CNN}

\author{Georgia Gkioxari\\
UC Berkeley\\
{\tt\small gkioxari@eecs.berkeley.edu}
\and
Ross Girshick\\
Microsoft Research\\
{\tt\small rbg@microsoft.com}
\and
Jitendra Malik\\
UC Berkeley\\
{\tt\small malik@eecs.berkeley.edu}
}

\maketitle

\begin{abstract}
 There are multiple cues in an image which reveal what action a person is performing. For example, a jogger has a pose that is characteristic for jogging, but the scene (\eg road, trail) and the presence of other joggers can be an additional source of information. In this work, we exploit the simple observation that actions are accompanied by contextual cues to build a strong action recognition system. We adapt RCNN to use more than one region for classification while still maintaining the ability to localize the action. We call our system R$^*$CNN. The action-specific models and the feature maps are trained jointly, allowing for action specific representations to emerge. R$^*$CNN achieves 90.2\% mean AP on the PASAL VOC Action dataset, outperforming all other approaches in the field by a significant margin. Last, we show that R$^*$CNN is not limited to action recognition. In particular, R$^*$CNN can also be used to tackle fine-grained tasks such as attribute classification. We validate this claim by reporting state-of-the-art performance on the Berkeley Attributes of People dataset.\footnote{Source code and models are available at \url{https://github.com/gkioxari/RstarCNN}}
\end{abstract}

\input{sections/intro}
\input{sections/related}
\input{sections/approach}
\input{sections/results}

\section*{Conclusion}
We introduce a simple yet effective approach for action recognition. We adapt RCNN to use more than one region in order to make a prediction, based on the simple observation that contextual cues are significant when deciding what action a person is performing. We call our system \emph{R$^*$CNN}. In our setting, both features and models are learnt jointly, allowing for action-specific representations to emerge. R$^*$CNN outperforms all published approaches on two datasets. More interestingly, the auxiliary information selected by R$^*$CNN for prediction captures different contextual modes depending on the instance in question. 

 R$^*$CNN is not limited to action recognition. We show that R$^*$CNN can be used successfully for tasks such as attribute classification. Our visualizations show that the secondary regions capture the region relevant to the attribute considered.
 
 \section*{Acknowledgments}
 This work was supported by the Intel Visual Computing Center and the ONR SMARTS MURI N000140911051. The GPUs used in this research were generously donated by
the NVIDIA Corporation.

{\small
\bibliographystyle{ieee}
\bibliography{refs}
}

\end{document}

%% file: sections/intro.tex
\section{Introduction}
\seclabel{intro}

Consider \figref{fig:Fig1} (a). How do we know that the person highlighted with the red box is working on a computer? Could it be that the computer is visible in the image, is it that the person in question has a very specific pose or is it that he is sitting in an office environment? Likewise, how do we know that the person in \figref{fig:Fig1} (b) is running? Is it the running-specific pose of her arms and legs or do the scene and the other people nearby also convey the action?

For the task of action recognition from still images, the pose of the person in question, the identity of the objects surrounding them and the way they interact with those objects and the scene are vital cues. In this work, our objective is to use all available cues to perform activity recognition.

Formally, we adapt the Region-based Convolutional Network method (RCNN) \cite{girshick2014rcnn} to use more than one region when making a prediction. We call our method \emph{R$^*$CNN}. In R$^*$CNN, we have a primary region that contains the person in question and a secondary region that automatically discovers contextual cues.

\begin{figure}
\begin{center}
  \includegraphics[width=1.0\linewidth]{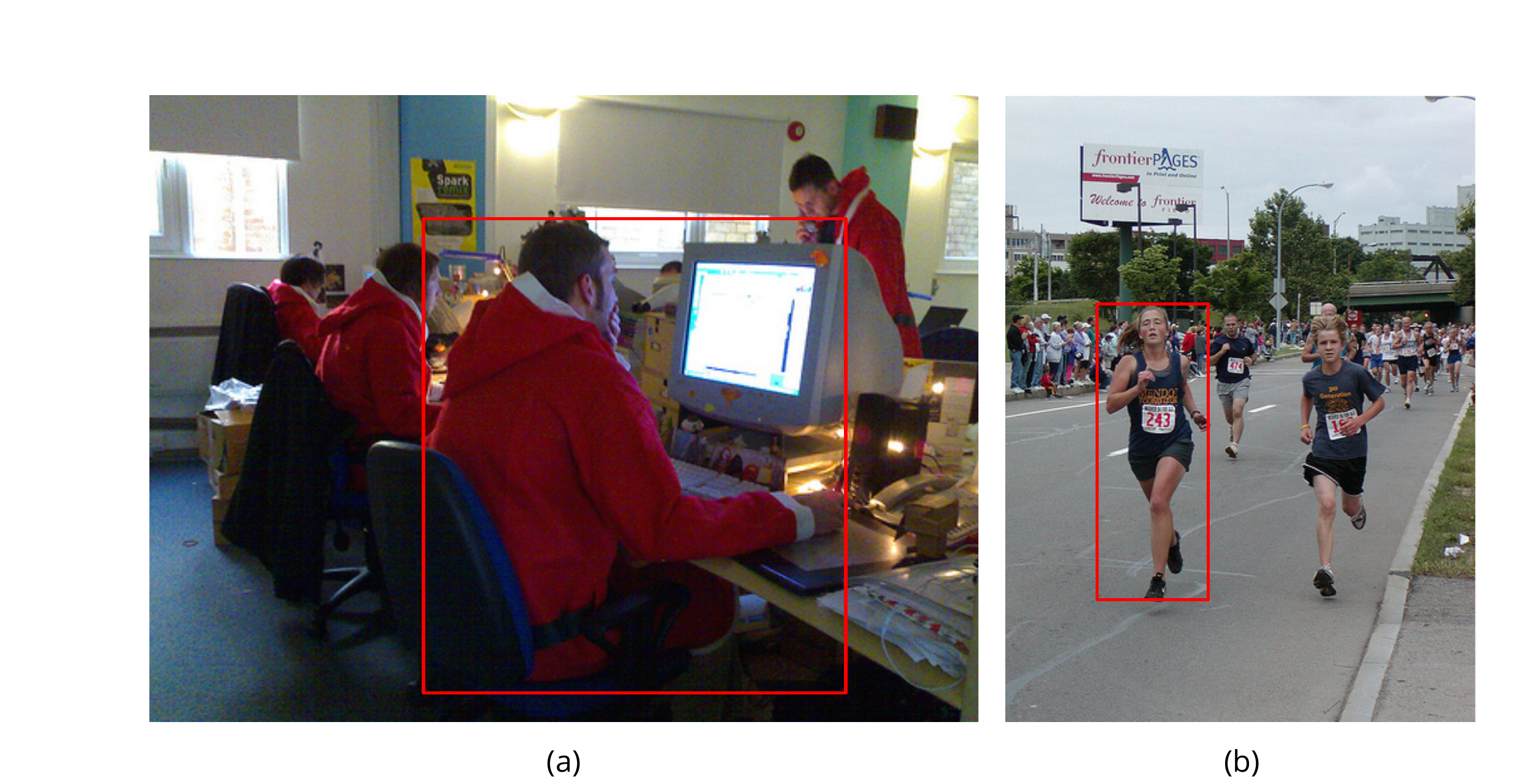}
\end{center}
\caption{Examples of people performing actions.}
   \figlabel{fig:Fig1}
\end{figure}

\begin{figure*}[t!]
\begin{center}
  \includegraphics[width=1.0\linewidth,trim=0 4em 0 0,clip=true]{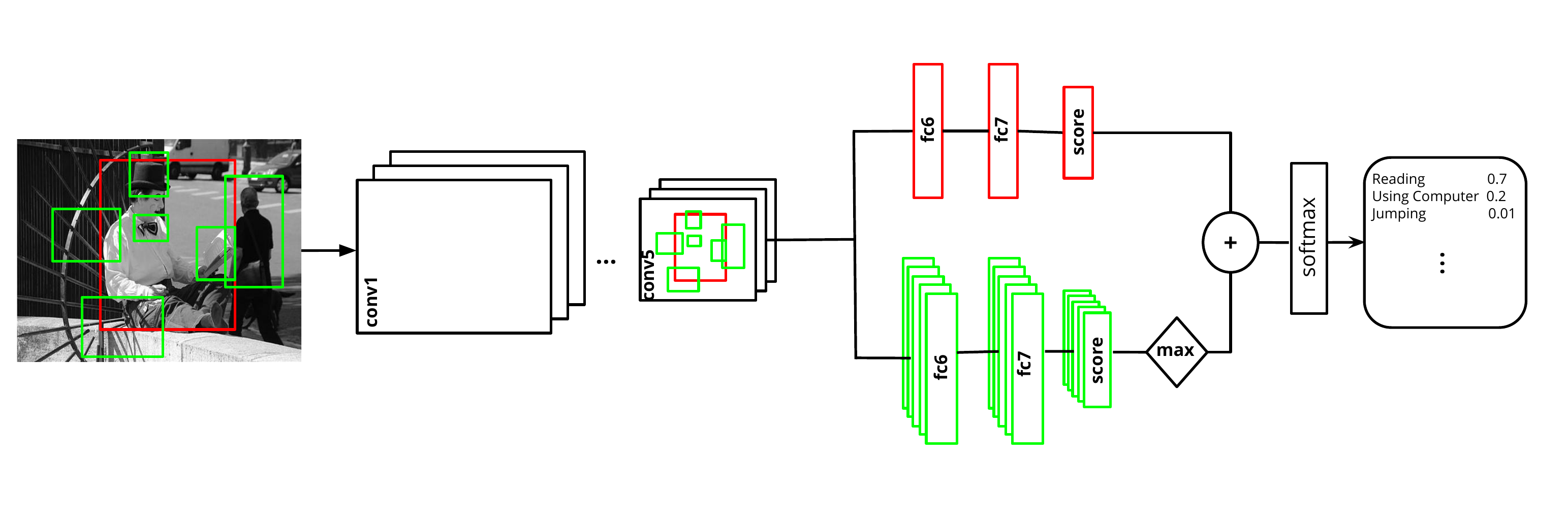}
\end{center}
\caption{Schematic overview of our approach. Given image $I$, we select the primary region to be the bounding box containing the person (\textcolor{red}{red box}) while region proposals define the set of candidate secondary regions (\textcolor{green}{green boxes}). For each action $\alpha$, the most informative secondary region is selected (\emph{max} operation) and its score is added to the primary. The \emph{softmax} operation transforms scores into probabilities and forms the final prediction.}
   \figlabel{fig:Overview}
\end{figure*}

How do we select the secondary region? In other words, how to we decide which region contains information about the action being performed? Inspired by multiple-instance learning (MIL) \cite{Viola05, Maron98} and Latent SVM \cite{lsvm-pami}, if $I$ is an image and $r$ is a region in $I$ containing the target person, we define the $\textrm{score}$ of action $\alpha$ as

\begin{equation}
  \textrm{score}(\alpha ; I,r) = \vecwp \cdot \bphi(r ; I) + \max_{s \in R(r ; I)} \vecws \cdot \bphi(s ; I),
\eqlabel{eq:MIL}
\end{equation}
where $\bphi(r ; I)$ is a vector of features extracted from region $r$ in $I$, while $\vecwp$ and $\vecws$ are the primary and secondary weights for action $\alpha$ respectively. $R(r ; I)$ defines the set of candidates for the secondary region. For example, $R(r ; I)$ could be the set of regions in the proximity of $r$, or even the whole set of regions in $I$. Given scores for each action, we use a softmax to compute the probability that the person in $r$ is performing action $\alpha$:

\begin{equation}
  P(\alpha | I, r) = \frac{\exp(\textrm{score}(\alpha ; I, r) )}{\sum_{\alpha' \in A} \exp(\textrm{score}(\alpha' ; I, r) )}.
\eqlabel{eq:Softmax}
\end{equation}

The feature representation $\bphi(\cdot)$ and the weight vectors $\vecwp$ and $\vecws$ in \eqref{eq:MIL} are learned \emph{jointly} for all actions $\alpha \in A$ using a CNN trained with stochastic gradient descent (SGD). We build on the Fast RCNN implementation \cite{FastRCNN}, which efficiently processes a large number of regions per image. \figref{fig:Overview} shows the architecture of our network.

We quantify the performance of R$^*$CNN for action recognition using two datasets: PASCAL VOC Actions \cite{PASCAL-ijcv} and the MPII Human Pose dataset \cite{andriluka14cvpr}.
On PASCAL VOC, R$^*$CNN yields 90.2\% mean AP, improving the previous state-of-the-art approach \cite{Simonyan2015} by 6 percentage points, according to the leaderboard \cite{leaderboard}. We visualize the selected secondary regions in \figref{fig:VOC_test} and show that indeed the secondary models learn to pick auxiliary cues as desired.
On the larger MPII dataset, R$^*$CNN yields 26.7\% mean AP, compared to 5.5\% mean AP achieved by the best performing approach, as reported by \cite{mpii-action}, which uses holistic \cite{wang2013} and pose-specific features along with motion cues.

In addition to the task of action recognition, we show that R$^*$CNN can successfully be used for fine-grained tasks. We experiment with the task of attribute recognition and achieve state-of-the-art performance on the Berkeley Attributes of People dataset \cite{BourdevAttributesICCV11}. Our visualizations in \figref{fig:Attributes} show that the secondary regions capture the parts specific to the attribute class being considered.

%% file: sections/related.tex
\section{Related Work}
\seclabel{related}

\paragraph{Action recognition.} There is a variety of work in the field of action recognition in static images. The majority of the approaches use holistic cues, by extracting features on the person bounding box and combining them with contextual cues from the whole image and object models.

Maji \etal \cite{MajiActionCVPR11} train action specific poselets and for each instance create a poselet activation vector that is classified using SVMs. They capture contextual cues in two ways: they explicitly detect objects using pre-trained models for the \emph{bicycle, motorbike, horse} and \emph{tvmonitor} categories and exploit knowledge of actions of other people in the image.
Hoai \etal \cite{Hoai-BMVC14} use body-part detectors and align them with respect to the parts of a similar instance, thus aligning their feature descriptors. They combine the part based features with object detection scores and train non-linear SVMs.  
Khosla \etal \cite{CVPR11_0254} densely sample image regions at arbitrary locations and scales with reference to the ground-truth region. They train a random forest classifier to discriminate between different actions. 
Prest \etal \cite{prest2012} learn human-object interactions using only action labels. They localize the action object by finding recurring patterns on images of actions and then capture their relative spatial relations.
The aforementioned approaches are based on hand-engineered features such as HOG \cite{Dalal05} and SIFT \cite{SIFT}.

CNNs achieve state-of-the-art performance on handwritten digit classification \cite{lecun-89e}, and have recently been applied to various tasks in computer vision such as image classification \cite{krizhevsky2012imagenet, Simonyan2015} and object detection \cite{girshick2014rcnn} with impressive results. 
For the task of action recognition, Oquab \etal \cite{Oquab14} use a CNN on ground-truth boxes for the task of action classification, but observe a small gain in performance compared to previous methods. 
Hoai \cite{Hoai-BMVC14-RMP} uses a geometrical distribution of regions placed in the image and in the ground-truth box and weights their scores to make a single prediction, using fc7 features from a network trained on the ImageNet-1k dataset \cite{ILSVRC12}. 
Gkioxari \etal \cite{deepparts} train body part detectors (\emph{head, torso, legs}) on pool5 features in a sliding-window manner and combine them with the ground-truth box to jointly train a CNN.

Our work is different than the above mentioned approaches in the following ways. 
We use bottom up region proposals \cite{UijlingsIJCV2013} as candidates for secondary regions, instead of anchoring regions of specific aspect ratios and at specific locations in the image, and without relying on the reference provided by the ground-truth bounding box. Region proposals have been shown to be effective object candidates allowing for detection of objects irrespective of occlusion and viewpoint. 
We jointly learn the feature maps and the weights of the scoring models, allowing for action specific representations to emerge. These representations might refer to human-object relations, human-scene relations and human-human relations. This approach is contrary to work that predefines the relations to be captured or that makes use of hand-engineered features, or features from networks trained for different tasks. We allow the classifier to pick the most informative secondary region for the task at hand. As we show in \secref{results}, the selected secondary region is instance specific and can be an object (\eg, cell phone), a part of the scene (\eg, nearby bicycles), the whole scene, or part of the human body.

\paragraph{Scene and Context.} The scene and its role in vision and perception have been studied for a long time. Biederman \etal \cite{biederman1982scene} identify five classes of relationships (presence, position, size, support and interposition) between an object and its setting and conduct experiments to measure how well humans identify objects when those relationships are violated. They found that the ability to recognize objects is much weaker and it becomes worse as violations become more severe.  More recently, Oliva and Torralba \cite{oliva2007} study the contextual associations of objects with their scene and link various forms of context cues with computer vision.

\paragraph{Multiple-Instance Learning.} Multiple instance learning (MIL) provides a framework for training models when full supervision is not available at train time. Instead of accurate annotations, the data forms bags, with a positive or a negative label \cite{Maron98}. There is a lot of work on MIL for computer vision tasks.  For object detection, Viola \etal \cite{Viola05} use MIL and boosting to obtain face detectors when ground truth object face locations are not accurately provided at train time. More recently, Song \etal \cite{song14slsvm} use MIL to localize objects with binary image-level labels (is the object present in the image or not). For the task of image classification, Oquab \etal \cite{oquab2015} modify the CNN architecture \cite{krizhevsky2012imagenet}, which divides the image into equal sized regions and combines their scores via a final max pooling layer to classify the whole image. Fang \etal \cite{haoCVPR15} follow a similar technique to localize concepts useful for image caption generation.

In this work, we treat the secondary region for each training example as an unknown latent variable.
During training, each time an example is sampled, the forward pass of the CNN infers the current value of this latent variable through a max operation.
This is analogous to latent parts locations and component models in DPM \cite{lsvm-pami}.
However, here we perform end-to-end optimization with an online algorithm (SGD), instead of optimizing a Latent SVM.

%% file: sections/approach.tex
\section{Implementation}
\seclabel{approach}

\figref{fig:Overview} shows the architecture of our network. Given an image $I$, we select the primary region to be the bounding box containing the person (knowledge of this box is given at test time in all action datasets). Bottom up region proposals form the set of candidate secondary regions. For each action $\alpha$, the most informative region is selected through the \emph{max} operation and its score is added to the primary (\eqref{eq:MIL}). The \emph{softmax} operation transforms scores into estimated posterior probabilities (\eqref{eq:Softmax}), which are used to predict action labels.

\subsection{R$^*$CNN}
We build on Fast RCNN (FRCN) \cite{FastRCNN}. In FRCN, the input image is upsampled and passed through the convolutional layers. An adaptive max pooling layer takes as input the output of the last convolutional layer and a list of regions of interest (ROIs).
It outputs a feature map of fixed size (\eg $7\times7$ for the 16-layer CNN by \cite{Simonyan2015}) specific to each ROI. The ROI-pooled features are subsequently passed through the fully connected layers to make the final prediction. This implementation is efficient, since the computationally intense convolutions are performed at an image-level and are subsequently being reused by the ROI-specific operations.

The test-time operation of FRCN is similar to SPPnet \cite{sppnets}. However, the training algorithm is different and enables fine-tuning all network layers, not just those above the final ROI pooling layer, as in \cite{sppnets}.
This property is important for maximum classification accuracy with very deep networks.

In our implementation, we extend the FRCN pipeline. Each primary region $r$ of an image $I$ predicts a score for each action $\alpha \in A$ (top stream in \figref{fig:Overview}). At the same time, each region within the set of candidate secondary regions $R(r ; I)$ independently makes a prediction.
These scores are combined, for each primary region $r$, by a \emph{max} operation over $r$'s candidate regions (bottom stream in \figref{fig:Overview}).

We define the set of candidate secondary regions $R(r ; I)$ as
\begin{equation}
R(r ; I) = \{ s \in S(I) : \textrm{overlap}(s,r) \in [l,u] \} ,
\eqlabel{eq:R}
\end{equation}
where $S(I)$ is the set of region proposals for image $I$. In our experiments, we use Selective Search \cite{UijlingsIJCV2013}. The lower and upper bounds for the overlap, which here is defined as the intersection over union between the boxes, defines the set of the regions that are considered as secondary for each primary region. For example, if $l=0$ and $u=1$ then $R(r ; I) = S(I)$, for each $r$, meaning that all bottom up proposals are candidates for secondary regions.

\subsection{Learning}

We train R$^*$CNN with stochastic gradient descent (SGD) using backpropagation. We adopt the 16-layer network architecture from \cite{Simonyan2015}, which has been shown to perform well for image classification and object detection. 

During training, we minimize the log loss of the predictions. If $P(\alpha~|~I, r)$ is the softmax probability that action $\alpha$ is performed in region $r$ in image $I$ computed by \eqref{eq:Softmax}, then the loss over a batch of training examples $B = \{ I_i, r_i, l_i\}_{i=1}^{M}$ is given by

\begin{equation}
  \textrm{loss}(B) = -\frac{1}{M}\sum_{i=1}^{M} \log P(\alpha = l_i~|~I_i, r_i),
\eqlabel{eq:Loss}
\end{equation}
where $l_i$ is the true label of example $r_i$ in image $I_i$.

Rather than limiting training to the ground-truth person locations, we use all regions that overlap more than $0.5$ with a ground-truth box. This condition serves as a form of data augmentation. For every primary region, we randomly select $N$ regions from the set of candidate secondary regions. $N$ is a function of the GPU memory limit (we use a Nvidia K40 GPU) and the batch size. 

We fine-tune our network starting with a model trained on ImageNet-1K for the image classification task. We tie the weights of the fully connected primary and secondary layers (\emph{fc6, fc7}), but not for the final scoring models. We set the learning rate to $0.0001$, the batch size to $30$ and consider 2 images per batch. We pick $N=10$ and train for 10K iterations. 
Larger learning rates prevented fine-tuning from converging.

Due to the architecture of our network, most computation time is spent during the initial convolutions, which happen over the whole image. Computation does not scale much with the number of boxes, contrary to the original implementation of RCNN \cite{girshick2014rcnn}. Training takes 1s per iteration, while testing takes 0.4s per image.

%% file: sections/results.tex
\section{Results}
\seclabel{results}

We demonstrate the effectiveness of R$^*$CNN on action recognition from static images on the PASCAL VOC Actions dataset \cite{PASCAL-ijcv}, the MPII Human Pose dataset \cite{andriluka14cvpr} and the Stanford 40 Actions dataset \cite{yao2011human}.

\subsection{PASCAL VOC Action}

The PASCAL VOC Action dataset consists of 10 different actions, \emph{Jumping, Phoning, Playing Instrument, Reading, Riding Bike, Riding Horse, Running, Taking Photo, Using Computer, Walking} as well as examples of people not performing some of the above action, which are marked as \emph{Other}. The ground-truth boxes containing the people are provided both at train and test time. During test time, for every example we estimate probabilities for all actions and compute AP. 

\subsubsection{Control Experiments}

We experiment with variants of our system to show the effectiveness of R$^*$CNN.

\begin{itemize}

\item {\bf{RCNN}}. As a baseline approach we train Fast R-CNN for the task of action classification. This network exploits only the information provided from the primary region, which is defined as the ground-truth region. 

\item {\bf{Random-RCNN}}. We use the ground-truth box as a primary region and a box randomly selected from the secondary regions. We train a network for this task similar to R$^*$CNN with the \emph{max} operation replaced by \emph{rand}

\item {\bf{Scene-RCNN}}. We use the ground-truth box as the primary region and the whole image as the secondary. We jointly train a network for this task, similar to R$^*$CNN, where the secondary model learns action specific weights solely from the scene (no \emph{max} operation is performed in this case)

\item {\bf{R$^*$CNN $(l,u)$}}. We experiment with various combinations of values for the only free parameters of our pipeline, namely the bounds $(l,u)$ of the overlaps used when defining the secondary regions $R(r; I)$, where $r$ is the primary region

\item {\bf{R$^*$CNN $(l,u,n_S)$}}. In this setting, we use $n_S>1$ secondary regions instead of one. The secondary regions are selected in a greedy manner. First we select the secondary region $s_1$ exactly as in R$^*$CNN. The $i$-th secondary region $s_i$ is selected via the \emph{max} operation from the set $R(r ; I) \cap R(s_1 ; I) \cap ... \cap R(s_{i-1} ; I)$, where $r$ is the primary region. 

\end{itemize}

The Random- and Scene- settings show the value of selecting the most informative region, rather than forcing the secondary region to be the scene or a region selected at random.

\tableref{tab:Action_val} shows the performance of all the variants on the val set of the PASCAL VOC Actions. Our experiments show that R$^*$CNN performs better across all categories. In particular, \emph{Phoning}, \emph{Reading}, \emph{Taking Photo} perform significantly better than the baseline approach and Scene-RCNN. \emph{Riding Bike}, \emph{Riding Horse} and \emph{Running} show the smallest improvement, probably due to scene bias of the images containing those actions. Another interesting observation is that our approach is not sensitive to the bounds of overlap $(l,u)$. R$^*$CNN is able to perform very well even for the unconstrained setting where all regions are allowed to be picked by the secondary model, $(l=0, u=1)$. In our basic R$^*$CNN setting, we use one secondary region. However, one region might not be able to capture all the modes of contextual cues present in the image. Therefore, we extend R$^*$CNN to include $n_S$ secondary regions. Our experiments show that for $n_S=2$ the performance is the same as with R$^*$CNN for the optimal set of parameters of $(l=0.2, u=0.75)$.

\begin{table*}
\centering
\renewcommand{\arraystretch}{1.2}
\renewcommand{\tabcolsep}{1.2mm}
\resizebox{\linewidth}{!}{
\begin{tabular}{@{}l|r*{9}{c}|cc@{}}
AP (\%)  & Jumping  & Phoning & Playing Instrument & Reading & Riding Bike & Riding Horse & Running & Taking Photo & Using Computer & Walking & mAP \\
\hline
RCNN                               & 88.7 & 72.6 & 92.6 & 74.0 & 96.1 & 96.9 & 86.1 & 83.3 & 87.0 & 71.5 & 84.9  \\
Random-RCNN                & 89.1 & 72.7 & 92.9 & 74.4 & 96.1 & 97.2 & 85.0 & 84.2 & 87.5 & 70.4 & 85.0  \\     
Scene-RCNN                   & 88.9 & 72.5 & 93.4 & 75.0 & 95.6 & 98.1 &  88.6 & 83.2 & 90.4 & 71.5 & 85.7 \\     
R$^*$CNN (0.0, 0.5)        & 89.1 & 80.0 & \bf{95.6} & 81.0 & \bf{97.3} & 98.7 & 85.5 & \bf{85.6} & 93.4 & 71.5 & 87.8 \\
R$^*$CNN (0.2, 0.5)        & 88.1 & 75.4 & 94.2 & 80.1 & 95.9 & 97.9 & 85.6 & 84.5 & 92.3 & \bf{71.6} & 86.6 \\
R$^*$CNN (0.0, 1.0)        & \bf{89.2} & 77.2 & 94.9 & \bf{83.7} & 96.7 & \bf{98.6} & 87.0 & 84.8 & 93.6 & 70.1 & 87.6 \\
R$^*$CNN (0.2, 0.75)      & 88.9 & 79.9 & 95.1 & 82.2 & 96.1 & 97.8 & \bf{87.9} & 85.3 & 94.0 & 71.5 & \bf{87.9} \\
R$^*$CNN (0.2, 0.75, 2)  & 87.7 & \bf{80.1} & 94.8 & 81.1 & 95.5 & 97.2 & 87.0 & 84.7 & \bf{94.6} & 70.1 & 87.3 
\end{tabular}
}
 \vspace{0.1em}
 \caption{AP on the PASCAL VOC Action 2012 val set. \emph{RCNN} is the baseline approach, with the ground-truth region being the primary region. \emph{Random-RCNN} is a network trained with primary the ground-truth region and secondary a random region. \emph{Scene-RCNN} is a network trained with primary the ground-truth region and secondary the whole image. \emph{R$^*$CNN $(l, u)$} is our system where $l,u$ define the lower and upper bounds of the allowed overlap of the secondary region with the ground truth. \emph{R$^*$CNN $(l, u, n_S)$} is a variant in which $n_S$ secondary regions are used, instead of one.}
\tablelabel{tab:Action_val}
\end{table*}

\begin{table*}
\centering
\renewcommand{\arraystretch}{1.2}
\renewcommand{\tabcolsep}{1.2mm}
\resizebox{\linewidth}{!}{
\begin{tabular}{@{}l|c|r*{9}{c}|cc@{}}
AP (\%)  & CNN layers & Jumping  & Phoning & Playing Instrument & Reading & Riding Bike & Riding Horse & Running & Taking Photo & Using Computer & Walking & mAP \\
\hline
Oquab \etal \cite{Oquab14}                                       & 8     & 74.8 & 46.0 & 75.6 & 45.3 & 93.5 & 95.0 & 86.5 & 49.3 & 66.7 & 69.5 & 70.2   \\
Hoai \cite{Hoai-BMVC14-RMP}                                & 8     &  82.3 & 52.9 & 84.3 & 53.6 & 95.6 & 96.1 & 89.7 & 60.4 & 76.0 & 72.9 & 76.3\\ 
Gkioxari \etal \cite{deepparts}                                   & 16     &  84.7 & 67.8 & 91.0 & 66.6 & 96.6 & 97.2 & 90.2 & 76.0 & 83.4 & 71.6 & 82.6 \\       
Simonyan \& Zisserman \cite{Simonyan2015}         & 16 \& 19  &  89.3 & 71.3 & \bf{94.7} & 71.3 & \bf{97.1} & 98.2 & 90.2 & 73.3 & 88.5 & 66.4 & 84.0 \\
R$^*$CNN                                                                & 16  &  \bf{91.5} & \bf{84.4} & 93.6 & \bf{83.2} & 96.9 & \bf{98.4} & \bf{93.8} & \bf{85.9} & \bf{92.6} & \bf{81.8} & \bf{90.2}
\end{tabular}
}
 \vspace{0.1em}
\caption{AP on the PASCAL VOC Action 2012 test set. Oquab \etal \cite{Oquab14} train an 8-layer network on ground-truth boxes. Gkioxari \etal \cite{deepparts} use part detectors for \emph{head, torso, legs} and train a CNN. Hoai \cite{Hoai-BMVC14-RMP} uses an 8-layer network to extract fc7 features from regions at multiple locations and scales. Simonyan and Zisserman \cite{Simonyan2015} combine a 16-layer and a 19-layer network and train SVMs on fc7 features from the image and the ground-truth box. R$^*$CNN (with $(l=0.2, u = 0.75)$) outperforms all other approaches by a significant margin. }
\tablelabel{tab:Action_test}
\end{table*}

\subsubsection{Comparison with published results}

We compare R$^*$CNN to other approaches on the PASCAL VOC Action test set. \tableref{tab:Action_test} shows the results. 
Oquab \etal \cite{Oquab14} train an 8-layer network on ground-truth boxes. 
Gkioxari \etal \cite{deepparts} use part detectors for \emph{head, torso, legs} and train a CNN on the part regions and the ground-truth box. 
Hoai \cite{Hoai-BMVC14-RMP} uses an 8-layer network to extract fc7 features from regions at multiple locations and scales inside the image and and the box and accumulates their scores to get the final prediction. 
Simonyan and Zisserman \cite{Simonyan2015} combine a 16-layer and a 19-layer network and train SVMs on fc7 features from the image and the ground-truth box. R$^*$CNN (with $(l=0.2, u = 0.75)$) outperforms all other approaches by a substantial margin. 
R$^*$CNN seems to be performing significantly better for actions which involve small objects and action-specific pose appearance, such as \emph{Phoning}, \emph{Reading}, \emph{Taking Photo}, \emph{Walking}.

\subsubsection{Visualization of secondary regions}

\figref{fig:VOC_test} shows examples from the top predictions for each action on the test set. Each block corresponds to a different action. Red highlights the person to be classified while green the automatically selected secondary region. For actions \emph{Jumping}, \emph{Running} and \emph{Walking} the secondary region is focused either on body parts (\eg legs, arms) or on more instances surrounding the instance in question (\eg joggers). For \emph{Taking Photo}, \emph{Phoning}, \emph{Reading} and \emph{Playing Instrument} the secondary region focuses almost exclusively on the object and its interaction with the arms. For \emph{Riding Bike}, \emph{Riding Horse} and \emph{Using Computer} it focuses on the object, or the presence of similar instances and the scene. 

Interestingly, the secondary region seems to be picking different cues depending on the instance in question. For example in the case of \emph{Running}, the selected region might highlight the scene (\eg road), parts of the human body (\eg legs, arms) or a group of people performing the action, as shown in \figref{fig:VOC_test}.

\begin{figure*}
\begin{center}
  \includegraphics[width=0.9\linewidth]{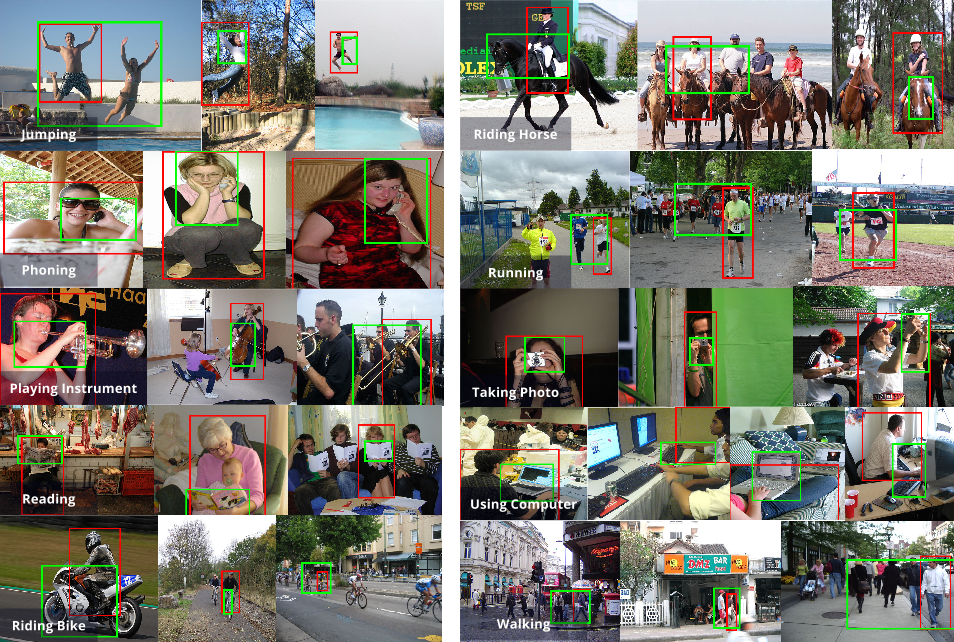}
\end{center}
\caption{Top predictions on the PASCAL VOC Action test set. The instance in question is shown with a \textcolor{red}{red box}, while the selected secondary region with a \textcolor{green}{green box}. The nature of the secondary regions depends on the action and the image itself. Even within the same action category, the most informative cue can vary.}
   \figlabel{fig:VOC_test}
\end{figure*}

\figref{fig:VOC_errs} shows erroneous predictions for each action on the val set (in descending score). Each block corresponds to a different action. The misclassified instance is shown in red and the corresponding secondary region with green. For \emph{Riding Bike} and \emph{Riding Horse}, which achieve a very high AP, the mistakes are of very low score.  For \emph{Jumping}, \emph{Phoning} and \emph{Using Computer} the mistakes occur due to confusions with instances of similar pose. In addition, for \emph{Playing Instrument} most of the misclassifications are people performing in concert venues, such as singers. For \emph{Taking Photo} and \emph{Playing Instrument} the presence of the object seems to be causing most misclassifications. For \emph{Running} and \emph{Walking} they seem to often get confused with each other as well as with standing people (an action which is not present explicitly in the dataset).

\begin{figure}
\begin{center}
  \includegraphics[width=0.9\linewidth]{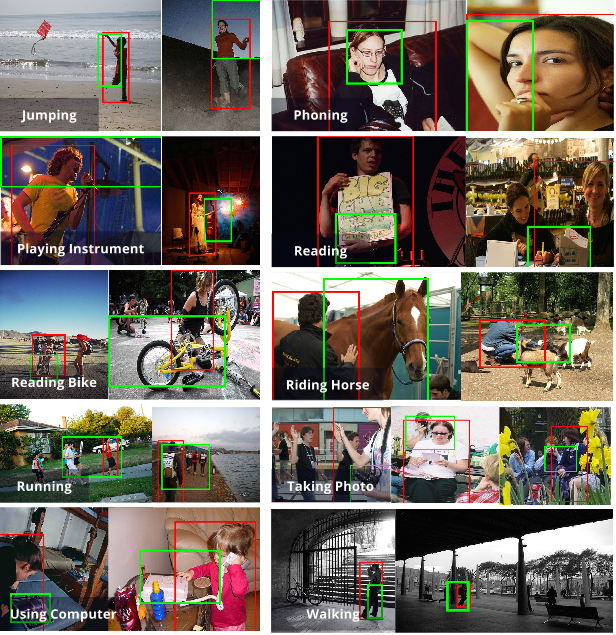}
\end{center}
\caption{Top mistakes on the PASCAL VOC Action val set. The misclassified instance is shown in \textcolor{red}{red}, while the selected secondary region in \textcolor{green}{green}.}
   \figlabel{fig:VOC_errs}
\end{figure}

\subsection{MPII Human Pose Dataset}

The MPII Human Pose dataset contains 400 actions and consists of approximately 40,000 instances and 24,000 images. The images are extracted from videos from YouTube. The training set consists of 15,200 images and 22,900 instances performing 393 actions. The number of positive training examples per category varies drastically \cite{mpii-action}. The amount of training data ranges from 3 to 476 instances, with an average of 60 positives per action. The annotations do not include a ground-truth bounding box explicitly, but provide a point (anywhere in the human body) and a rough scale of the human. This information can be used to extract a rough location of the instance, which is used as input in our algorithm. 

\subsubsection{R$^*$CNN vs. RCNN}

We split the training set into train and val sets. We make sure that frames of the same video belong to the same split to avoid overfitting. This results in 12,500 instances in train and 10,300 instances in val. We train the baseline RCNN network and R$^*$CNN. We pick $(l=0.2, u=0.5)$ due to the large number of region proposals generated by \cite{UijlingsIJCV2013} (on average 8,000 regions per image).

On the val set, RCNN achieves 16.5\% mean AP while R$^*$CNN achieves 21.7\% mean AP, across all actions. \figref{fig:MPII_val} shows the performance on MPII val for RCNN and R$^*$CNN. On the \textbf{left}, we show a scatter plot of the AP for all actions as a function of their training size. On the \textbf{right}, we show the mean AP across actions belonging to one out of three categories, depending on their training size. 

The performance reported in \figref{fig:MPII_val} is instance-specific. Namely, each instance is evaluated. One could evaluate the performance at the frame-level (as done in \cite{mpii-action}), \ie classify the frame and not the instance. We can generate frame-level predictions by assigning for each action the maximum score across instances in the frame. That yields 18.2\% mean AP for RCNN and 23\% mean AP for R$^*$CNN.

\begin{figure}
\begin{center}
  \includegraphics[width=1.0\linewidth]{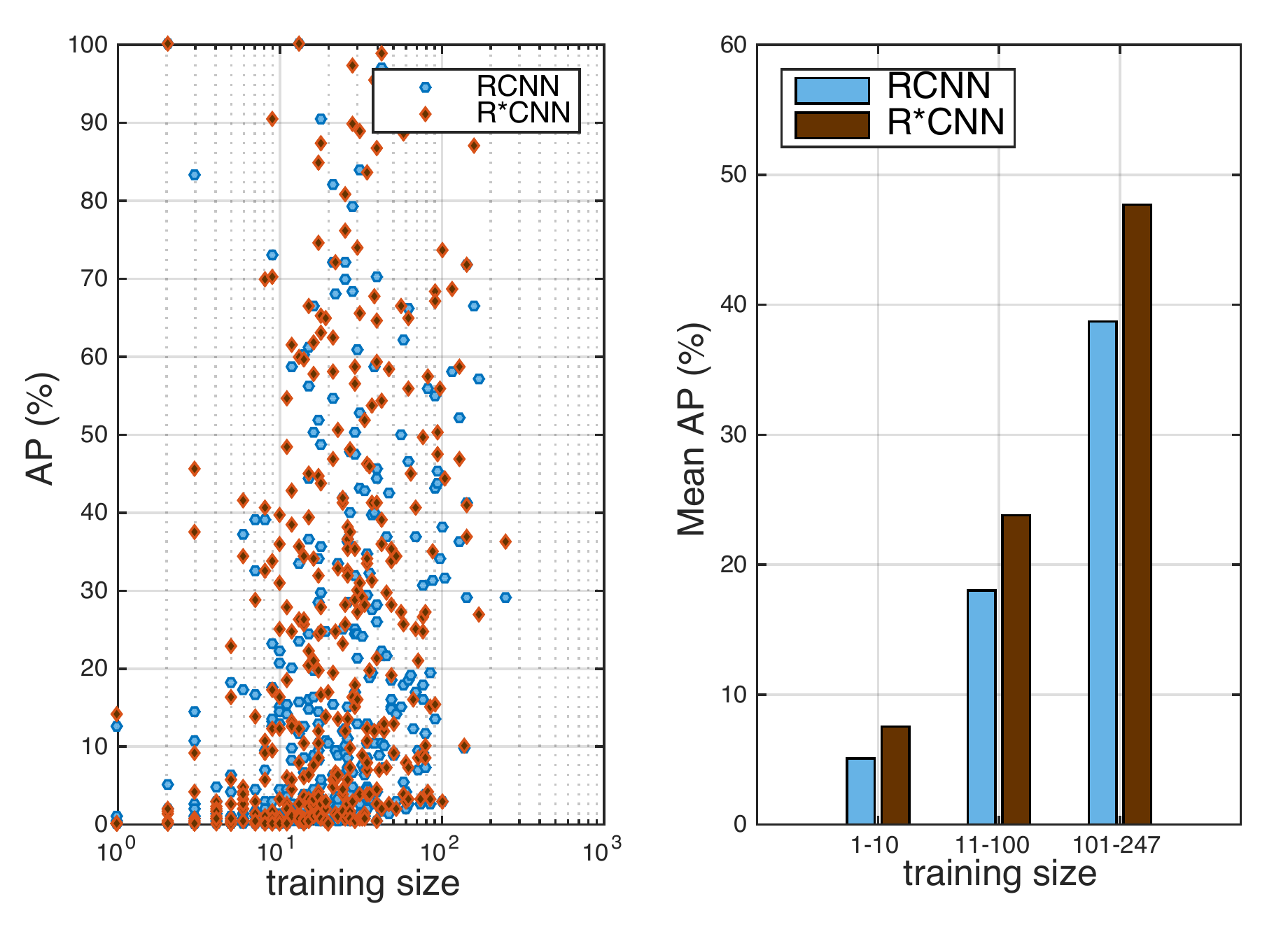}
\end{center}
\caption{Performance on MPII val for RCNN (\textcolor{ProcessBlue}{blue} ) and R$^*$CNN (\textcolor{Brown}{brown}). \textbf{Left:} AP (\%) for all actions as a function of their training size ($x$-axis). \textbf{Right:} Mean AP (\%) for three discrete ranges of training size  ($x$-axis).}
   \figlabel{fig:MPII_val}
\end{figure}

\subsubsection{Comparison with published results}

In \cite{mpii-action}, various approaches for action recognition are reported on the test set. All the approaches mentioned use motion features, by using frames in the temporal neighborhood of the frame in question. The authors test variants of Dense Trajectories (DT) \cite{wang2013} which they combine with pose specific features. The best performance on the test set is 5.5\% mean AP (frame-level) achieved by the DT combined with a pose specific approach. 

We evaluate R$^*$CNN on the test set\footnote{We sent our results to the authors of \cite{mpii-action} for evaluation since test annotations are not publicly available.} and achieve 26.7\% mAP for frame-level recognition. Our approach does not use motion, which is a strong cue for action recognition in video, and yet manages to outperform DT by a significant margin. Evaluation on the test set is performed only at the frame-level.

\figref{fig:MPII_test}  shows the mean AP across actions in a descending order of training size. This figure allows for a direct comparison with the published results, as shown in Figure 1(b) in \cite{mpii-action}.

\figref{fig:MPII_test_res} shows some results on the test set. We highlight the instance in question with red, and the secondary box with green. The boxes for the instances were derived from the point annotations (some point on the person) and the rough scale provided at train and test time. The predicted action label is overlaid in each image.

\begin{figure}
\begin{center}
  \includegraphics[width=0.9\linewidth]{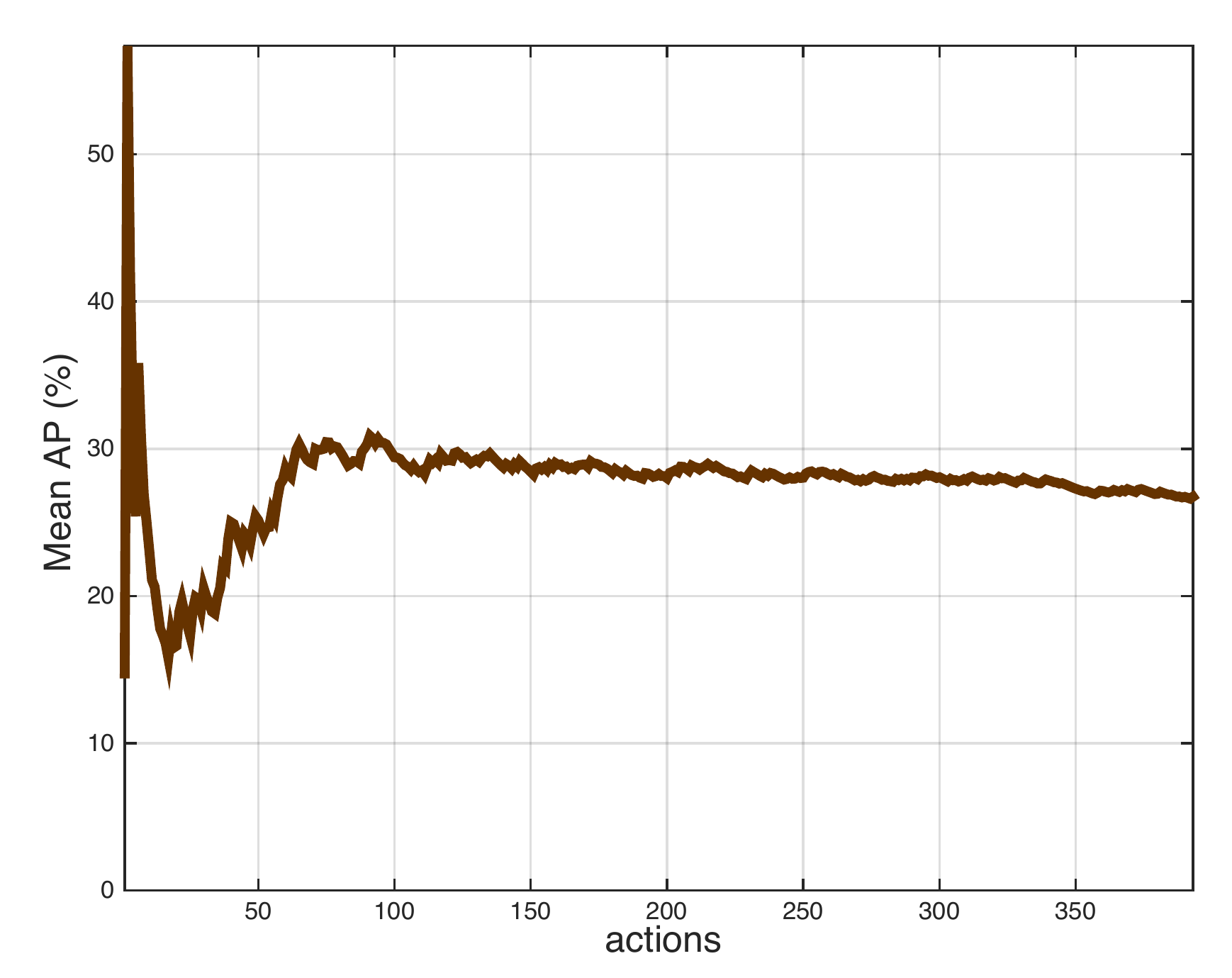}
\end{center}
\caption{Mean AP (\%) on MPII test for R$^*$CNN across actions in descending order of their training size. A direct comparison with published results, as shown in Figure 1(b) in \cite{mpii-action}, can be drawn.}
   \figlabel{fig:MPII_test}
\end{figure}

Even though R$^*$CNN outperforms DT, there is still need of movement to boost performance for many categories. For example, even though the MPII dataset has a many examples for actions such as \emph{Yoga}, \emph{Cooking or food preparation} and \emph{Video exercise workout}, R$^*$CNN performs badly on those categories (1.1\% mean AP). We believe that a hybrid approach which combines image and motion features, similar to \cite{simonyan2014, actiontubes}, would perform even better. 

\begin{figure*}
\begin{center}
  \includegraphics[width=0.42\linewidth]{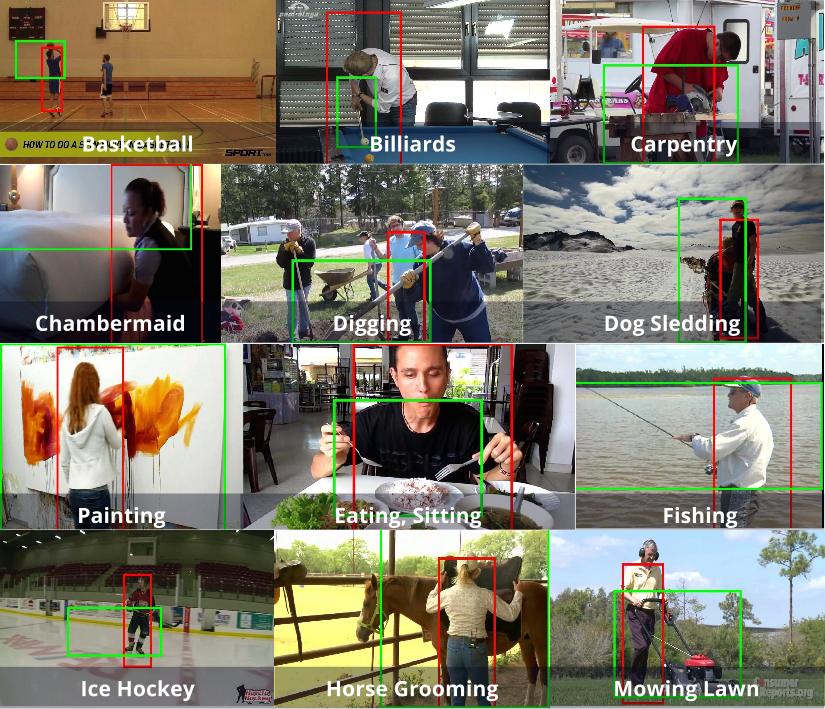}
  \includegraphics[width=0.45\linewidth]{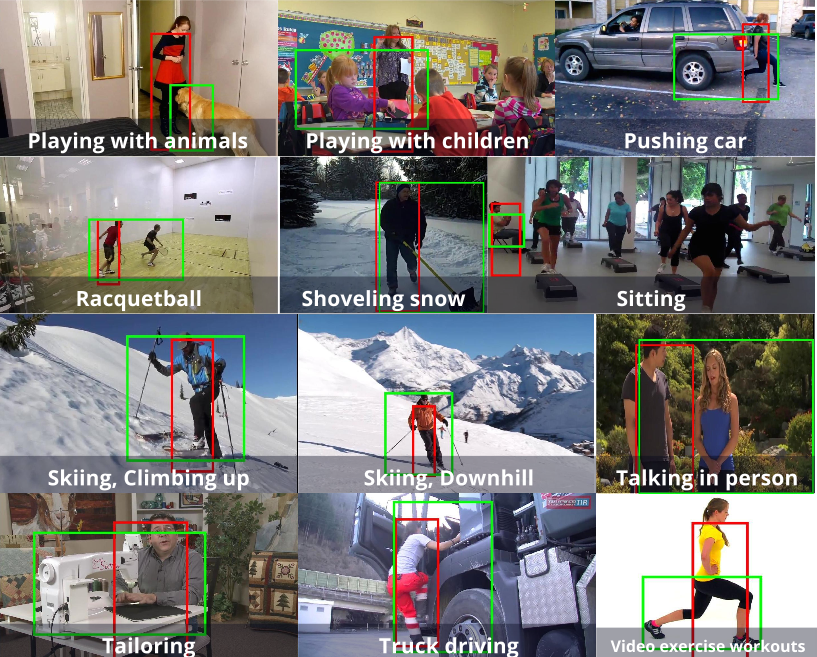}
\end{center}
\caption{Predictions on the MPII test set. We highlight the person in question with a \textcolor{red}{red} box, and the secondary region with a \textcolor{green}{green} box. The predicted action label is overlaid. }
   \figlabel{fig:MPII_test_res}
\end{figure*} 

\subsection{Stanford 40 Actions Dataset}

We run R$^*$CNN on the Stanford 40 Actions dataset \cite{yao2011human}. This dataset consists of 9532 images of people performing 40 different actions. The dataset is split in half to comprise the training and test split. Bounding boxes are provided for all people performing actions. R$^*$CNN achieves an average AP of 90.9\% on the test set, with performance varying from 70.5\% for \emph{texting message} to  100\% for \emph{playing violin}. \figref{fig:stanford40} shows the AP performance per action on the test set. Training code and models are publicly available.

\begin{figure}
\begin{center}
  \includegraphics[width=1.0\linewidth]{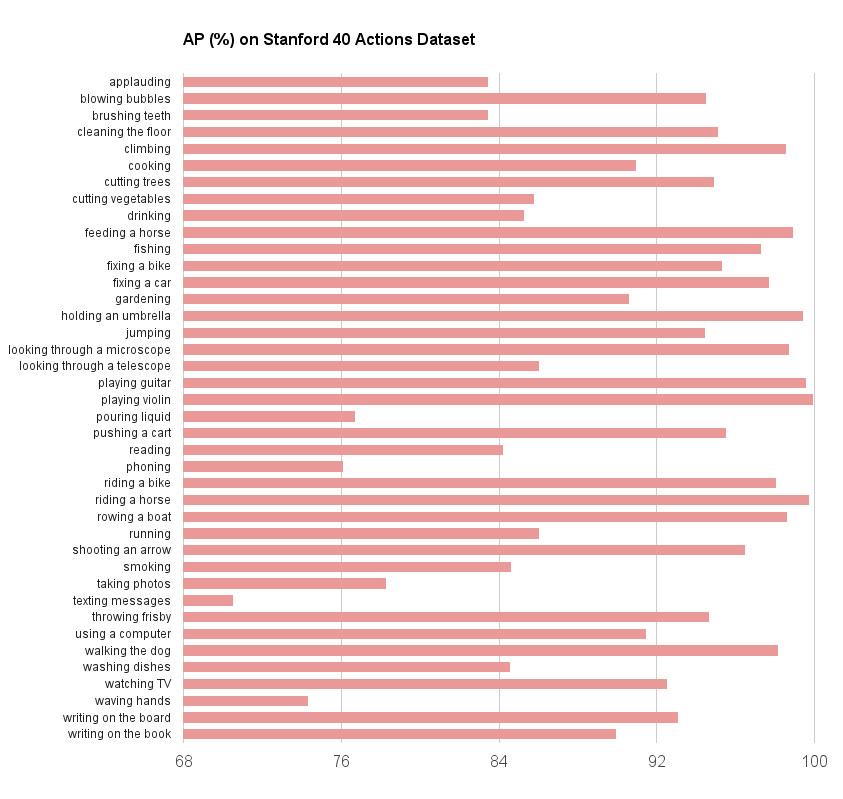}
\end{center}
\caption{AP (\%) of R$^*$CNN on the Stanford 40 dataset per action. Performance varies from 70.5\% for \emph{texting message} to 100\% for \emph{playing violin}. The average AP across all actions achieved by our model is 90.9\%.}
   \figlabel{fig:stanford40}
\end{figure}

\subsection{Attribute Classification}

\begin{table*}
\centering
\renewcommand{\arraystretch}{1.2}
\renewcommand{\tabcolsep}{1.2mm}
\resizebox{\linewidth}{!}{
\begin{tabular}{@{}l|c|r*{8}{c}|cc@{}}
AP (\%)  & CNN layers &  Is Male  & Has Long Hair & Has Glasses & Has Hat & Has T-Shirt & Has Long Sleeves & Has Shorts & Has Jeans & Has Long Pants & mAP \\
\hline
PANDA \cite{panda}                                          & 5 & 91.7 & 82.7 & 70.0 & 74.2 & 49.8 & 86.0 & 79.1 & 81.0 & 96.4 & 79.0   \\
Gkioxari \etal \cite{deepparts}                           & 16 & \bf{92.9} & \bf{90.1} & 77.7 & \bf{93.6} & 72.6 & \bf{93.2} & \bf{93.9} & \bf{92.1} & \bf{98.8} & \bf{89.5} \\   
RCNN                                                                & 16 &  91.8 & 88.9 & 81.0 & 90.4 & 73.1 & 90.4 & 88.6 & 88.9 & 97.6 & 87.8 \\     
R$^*$CNN                                                         & 16 & 92.8 & 88.9 & \bf{82.4} & 92.2 & \bf{74.8} & 91.2 & 92.9 & 89.4 & 97.9 & 89.2
\end{tabular}
}
 \vspace{0.1em}
\caption{AP on the Berkeley Attributes of People test set. PANDA \cite{panda} uses CNNs trained for each poselet type. Gkioxari \etal \cite{deepparts} detect parts and train a CNN jointly on the whole and the parts. RCNN is our baseline approach based on FRCN. Both RCNN and R$^*$CNN do not use any additional part annotations at training time. \cite{deepparts} and R$^*$CNN perform equally well, with the upside that R$^*$CNN does not need use keypoint annotations during training.}
\tablelabel{tab:Attributes_test}
\end{table*}

Finally, we show that R$^*$CNN can also be used for the task of attribute classification. On the Berkeley Attributes of People dataset \cite{BourdevAttributesICCV11}, which consists of images of people and their attributes, \eg \emph{wears hat}, \emph{is male} etc, we train R$^*$CNN as described above. The only difference is that our loss is no longer a log loss over softmax probabilities, but the cross entropy over independent logistics because attribute prediction is a multi-label task. \tableref{tab:Attributes_test} reports the performance in AP of our approach, as well as other competing methods. \figref{fig:Attributes} shows results on the test set. From the visualizations, the secondary regions learn to focus on the parts that are specific to the attribute being considered. For example, for the \emph{Has Long Sleeves} class, the secondary regions focus on the arms and torso of the instance in question, while for \emph{Has Hat} focus is on the face of the person.

\begin{figure}
\begin{center}
  \includegraphics[width=1.0\linewidth]{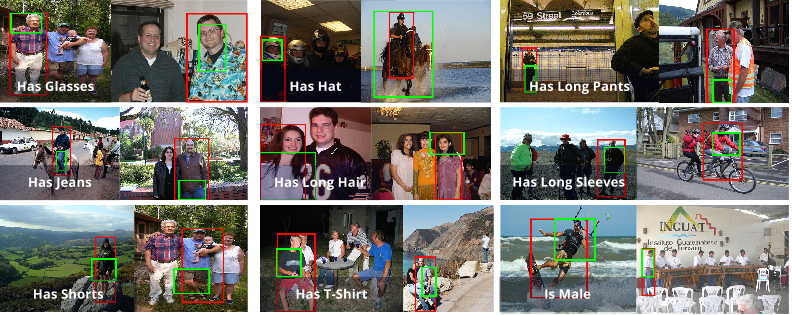}
\end{center}
\caption{Results on the Berkeley Attributes of People test set. We highlight the person in question with a \textcolor{red}{red} box, and the secondary region with a \textcolor{green}{green} box. The predicted attribute is overlaid. }
   \figlabel{fig:Attributes}
\end{figure}